4/9/03 v. 1.10.bit

# Blind normalization of speech from different channels

David N. Levin
Department of Radiology
University of Chicago
Chicago, Illinois 60637
d-levin@uchicago.edu

Running title: Normalization of speech from different channels

Abbreviated title: Blind normalization of speech from different channels

Authors address:
David N. Levin, M.D., Ph.D.
Dept. of Radiology, MC 2026
University of Chicago
5841 S. Maryland Ave.
Chicago, IL 60637

Email: d-levin@uchicago.edu
Tel.: 773-702-6511
Fax: 773-834-7610

## Abstract

We show how to construct a channel-independent representation of speech that has propagated through a noisy reverberant channel. This is done by blindly rescaling the cepstral time series by a non-linear function, with the form of this scale function being determined by previously encountered cepstra from that channel. The rescaled form of the time series is an invariant property of it in the following sense: it is unaffected if the time series is transformed by any time-independent invertible distortion. Because a linear channel with stationary noise and impulse response transforms cepstra in this way, the new technique can be used to remove the channel dependence of a cepstral time series. In experiments, the method achieved greater channel-independence than cepstral mean normalization, and it was comparable to the combination of cepstral mean normalization and spectral subtraction, despite the fact that no measurements of channel noise or reverberations were required (unlike spectral subtraction).

## I. INTRODUCTION

### A. The problem.

An ideal automatic speech recognition (ASR) system would be speaker-independent and channel-independent. After training during its manufacture, it would work "out of the box" to successfully recognize the utterances of different individuals speaking through a variety of channels. However, despite the steady progress of speech recognition technology in recent years, existing systems with large vocabularies are still sensitive to the nature of the acoustic environment and to the identity of the speaker[1-3]. For example, extensive retraining is often required if the acoustic channel is altered because the noise level changes, the speaker’s room or position changes, or the signal conduit changes (telephone vs. room speech). This paper considers the problem of designing speaker-dependent systems that are channel-independent. In other words, the objective is to create an ASR system that can accurately recognize a given speaker's utterances after they have propagated through any linear channel, once the system has been trained to recognize that speaker's speech from one linear channel. Ideally, the system would quickly adapt to changing channel conditions (e.g., to the changing noise and impulse response function of a moving speaker and/or microphone). This report describes a non-linear signal processing method that makes speech signals more channel-independent and that can be used in the "front end" of any ASR system.

**B. Conventional methods of achieving channel-independent ASR.**

In most commonly-encountered situations, the acoustic environment can be characterized in the time domain by a convolutive impulse response function and additive noise. In this case, the corrupted speech signal is parameterized by the filterbank outputs:

$$P_i = \int |X(f)|^2 |H(f)|^2 M_i(f) df + N_i \tag{1}$$

where $P_i$ is the power of the corrupted signal from the $i^{th}$ filterbank element, $|X(f)|^2$ is the power density of the channel's input (clean) signal, $|H(f)|^2$ is the power density of the channel's impulse response function, $M_i(f)$ is the profile of the $i^{th}$ filterbank element, and $N_i$ is the noise power from that element. This equation depends on the following approximations, which are commonly made and often work well in practice[1]: 1) the impulse response is small at time delays greater than the length of the spectral window; 2) the noise is not correlated with the speech. Notice that the noise term in Eq.(1) represents the noise power integrated over relatively wide filterbank elements (e.g., elements of a mel frequency filterbank). Therefore, to the extent that the underlying noise distribution is stationary and "white", this term is an average quantity with small frame-to-frame fluctuations.

Now, suppose that an ASR system was trained to recognize speech in one environment (e.g., clean speech) and it is now being used to analyze utterances from another channel (e.g., corrupted speech). If the channel transfer function $H$ is approximately constant over each filterbank element, it can be factored out of the integral in Eq.(1). Then, in the absence of noise, it simply has the effect of a translation in cepstral space, and cepstral mean normalization[4] (CMN) can be used to "subtract it out" in order to remove the effects of reverberations. However, if noise is present and/or the transfer function is narrow, there is a non-linear relationship between the cepstra from the two channels, and CMN is not as effective. Alternatively, reverberations can be compensated by accounting for the form of the impulse response after it has been measured by playing white noise, sine waves, or a chirp through the channel of interest[5]. However, this procedure can be impractical because such measurements may have to be repeated frequently in order to track a changing channel. Reverberations can also be combated with even more elaborate (and more cumbersome) methods involving microphone arrays[6].

The simplest way of accounting for noise is spectral subtraction, but this requires periodic noise power measurements[7]. Therefore, its implementation requires accurate discrimination between speech and no speech, which may require the help of the recognizer in the system's "back end". Wiener filtering can also be used to suppress noise, but this requires measurements of both the speech and noise power spectra[8].

The system's back end can be modified to incorporate the expected effects of a channel, but this can be computationally expensive. For example, in "multi-style training", the recognizer is trained on a database that contains speech samples from a variety of common channels[9]. In principle, this method has the disadvantage of "blurring" the statistical distributions of the recognizer, and, of course, it may perform poorly in the presence of an unanticipated channel. Alternatively, a clean speech model can be adapted to the channel of interest by using maximum likelihood linear regression[10] (MLLR) or by a parallel combination of clean speech and noise models[11].

However, this may entail a significant computational expense. Of course, one can attempt to fully retrain the recognizer for the channel of interest after it is first encountered. This is just the “brute force” approach that channel normalization techniques seek to avoid because it requires access to a full database of corrupted speech, and this must be measured or it must be synthesized from a database of clean speech after the channel's characteristics have been estimated. The computational burden is great, particularly if the system is meant to handle large vocabularies.

**C. The proposed method of channel normalization.**

Unlike existing ASR systems, humans perceive the information content of ordinary speech to be remarkably invariant in the presence of channel-dependent signal transformations. Yet there is no evidence that the speaker and listener exchange calibration data or that they measure the channel's impulse response and noise. Evidently, the speech signal is redundant in the sense that listeners blindly extract the same content from multiple acoustic signals that are transformed versions of one another. In earlier reports[12-16], the author showed how to design sensory devices that have this ability to recognize the underlying similarity of time-dependent signals differing by unknown transformations (linear or non-linear). In such devices, the signal is blindly rescaled by a non-linear function, with the form of this scale function being determined by previously encountered signal levels. The rescaled form of a signal time series is an invariant property of it in the following sense: it is unaffected if the time series is transformed by any time-independent invertible (one to one) distortion. In other words, the original time series and the transformed versions of it have the same rescaled form.

This is because a transformation's effect on the signal level at any time is compensated by its effect on the scale function. In earlier publications, this method was illustrated by applying it to analytic examples, simulated signals, acoustic waveforms of human speech, spectral time series of bird songs, and spectral time series of synthetic speech-like sounds[12-13, 15-16].

This approach is relevant to speech recognition for the following reason. After an utterance is passed through two different channels, the cepstral time series of the resulting output signals are related to one another by a non-linear transformation that characterizes the differences between the two channels. As shown in Section II, Eq.(1) implies the invertibility and time-independence of this transformation, as long as each channel's impulse response and noise distribution are stationary. Therefore, the same representation will result when the cepstral time series from either channel is blindly rescaled by the new signal processing method (Fig.1). Alternatively, the cepstral time series of utterances from channel #1 can be used to estimate the cepstral time series of the same utterances from channel #2 by first finding the invariant representation of the channel #1 signal and then synthesizing the channel #2 signal having the same invariant representation (dotted arrow in Fig. 1). This procedure can be performed as long as one has: 1) samples of a speaker's utterances from the two channels (possibly *different* utterances from each channel); 2) a few brief reference signals from each channel (each one a few milliseconds long), which represent the same input sounds and are used to define the origin and local orientation of each channel's scale function. Notice that the new method has the following advantages compared to conventional approaches to channel normalization: 1) it does not require explicit measurements of the channel's impulse response and noise; 2) it is a pure front end technology and avoids the computational demands of modifying or retraining the system's recognizer. Furthermore, the method can be run in an adaptive mode in order to account for changing channel conditions. This is done by using the most recent speech from the changing channel to continually update that channel's scale function.

The mathematical framework of the new method is described in Section II. Section III describes experiments in which the technique was used to normalize dimensionally-reduced speech data from different channels. The implications of these results and possible extensions of this work are discussed in Section IV.

## II. THEORY

In this section, we argue that a time-independent invertible transformation must relate the pair of cepstral time series, produced by the same utterance propagating through two time-independent channels. Then, we demonstrate how these two cepstral time series can be blindly rescaled so that they have the same representation. This rescaling process can be used to perform channel conversion: i.e., to modify the cepstral coefficients of an utterance from one channel so that they resemble those of the same utterance from another channel.

We make use of the embedding theorem that is well known in the field of non-linear dynamics[17]. This theorem states that almost every mapping from a *d*-dimensional space into a space of more than *2d* dimensions is invertible. Essentially, this is because so much "room" is provided by the "extra" dimensions of the higher dimensional space that the *d*-dimensional subspace, which is the range of the mapping, is very unlikely to self-intersect. Now, consider a speech signal that forms the input of any channel with stationary impulse response and noise. Because speech has 3-5 degrees of freedom[18-19] (S. Parthasarathy, AT&T Labs, private communication, 2001), the power spectra of this input signal lie in a 3-5-dimensional subspace within the space of all possible power spectra. For the linear channels described in Section I, the cepstral coefficients of the channel's output signal are time-independent functions of the input power spectra (Eq. (1)), and they lie in a 3-5-dimensional subspace within the space of all possible cepstra. The embedding theorem implies that this mapping is invertible, as long as we are using a sufficient number of cepstral coefficients (more than 6-10). Therefore, if the same input signal propagates through two different channels, the pair of output cepstral time series will be related by an invertible mapping, because each of them is invertibly related to the same time series of input power spectra. As is well known[1], this transformation between cepstra is quite non-linear if noise is present.

Let $x(t)$ ($x_k, k = 1,2,...,N$) be the time-dependent function that describes the trajectory of *N* cepstral coefficients of speech from a channel. In the following, we show how a special coordinate system (or scale) $s(x)$ is determined by a differential geometry that the speech trajectory imposes on the *x* manifold. Speech is invariantly represented in this coordinate system in the following sense: if its cepstral trajectory is subjected to any invertible transformation, the representation of the transformed trajectory in *its s* coordinate system is the same as the representation of the untransformed speech in *its* *s* coordinate system. To see how this comes about, consider a point *y* in a region of the *x* manifold that is densely sampled by the speech trajectory. Define $g^{kl}$ to be the average outer product of the time derivatives of the speech trajectory as it passes through a small neighborhood of *y*: $g^{kl} = \left\langle \frac{dx_k}{dt} \frac{dx_l}{dt} \right\rangle_{x(t) \sim y}$, where the bracket denotes the average over time. As long as this neighborhood contains *N* linearly independent time derivatives, $g^{kl}$ is positive definite, and its inverse $g_{kl}$ is well defined and positive definite. Under any change of coordinate systems, $x \to x' = x'(x)$, $\frac{dx}{dt}$ transforms as a contravariant vector. Therefore, $g^{kl}$ and $g_{kl}$ transform as a contravariant and covariant tensors, respectively. This means that $g_{kl}$ can be taken to define a metric on the *x* manifold, and a coordinate-independent process for moving (parallel transporting) vectors across the manifold can be derived from this metric by means of the methods of Riemannian geometry. For instance, the parallel transport process can be defined by means of an affine connection equal to the Christoffel symbol, which is composed of products of the metric's derivative and the inverse metric[20]. An attractive feature of this choice is that the parallel transport process is independent of the speaking rate. This is because $g^{kl}$ scales as the second power of this rate, $g_{kl}$ scales as its inverse second power, and the affine connection is unaffected. Now suppose that *N* linearly-independent "reference" vectors $h_a (a = 1,2,...,N)$ can be defined at a special "reference" point $x_0$ on the manifold. For example, in the experiments in Section III, the reference vectors were taken to be the average cepstral velocities of specified short segments of the speech trajectory as it passed through a specified neighborhood in cepstral space. Alternatively, as proposed in Section IV, the derivation of this reference information can be made almost completely automatic. The reference vectors can be parallel transported across the manifold to determine the *s* coordinates of any point *x*. Specifically, the point *x* can be assigned the coordinates *s* ($s_k, k = 1,2,...,N$), if it is reached by starting at $x_0$, then parallel transporting $h_1$ along itself $s_1$ times while simultaneously parallel transporting the other $h_a$ along the same path, then parallel transporting $h_2$ along itself $s_2$ times while simultaneously parallel transporting the other $h_a$ along the same path, …, and finally parallel transporting $h_N$ along itself $s_N$ times. In analogy to the problem of navigation, the reference vectors allow one to get one's "bearings" by establishing standard increments along "cardinal" directions at a certain point on the manifold. Once that is done, the parallel transport process can be used to carry those increments across the manifold in order to describe where other points are located with respect to the reference point. Notice that this parallel transport process is independent of what coordinate system is used on the cepstral (*x*) manifold[20]. Therefore, as long as the reference point/vectors can be identified in a coordinate-independent manner, the *s* representation of the speech trajectory will also be coordinate-independent. Because an invertible transformation of the trajectory is mathematically equivalent to a change of the manifold's coordinate system, this means that speech trajectories related by invertible transformations will have the same *s* representation. Recall that the embedding theorem implies the existence of an invertible mapping between the pair of speech trajectories of an utterance that propagated through two different channels. It follows that these trajectories have identical *s*-representations (Fig. 1). In principle, this representation can be used directly as channel-independent input of a recognizer.

However, in this report, we use this procedure to perform channel conversion: i.e., to modify the cepstral time series of speech from one channel (a corrupted channel) so that it resembles the cepstral time series of the same utterance from another (clean) channel (Fig. 1). Then, the converted cepstral coefficients can be fed into a conventional recognizer that has been trained on clean speech. To see how this is done, let $x(t)$ be the cepstral time series of an utterance from channel #1, and let $s(x)$ be the scale function derived from a speech sample from channel #1. Likewise, let $x'(t)$ be the cepstral trajectory of the same utterance from channel #2, and let $s'(x')$ be the scale function derived from the aforementioned speech sample after propagation through channel #2. In the previous paragraph, we showed that the rescaled representations of these two trajectories are the same: i.e., $s[x(t)] = s'[x'(t)]$. Therefore, the cepstral coefficients of the channel #1 speech can be found by mapping the *s*-representation of the channel #2 speech through the inverse of the scale function of the channel #1 speech: $x(t) = s^{-1}[s'[x'(t)]]$. Now, in the above discussion, it was assumed that the two scale functions were derived from identical speech samples that had propagated through the two channels. However, suppose that different utterances from the same speaker/channel combination always lead to the same metric and scale function. Then, the above channel conversion procedure can be performed even if different speech samples have been observed in the two channels. In other words, one can use the scale functions derived from different clean and corrupted speech samples to predict the cepstral coefficients of the clean versions of corrupted utterances. The success of the experiments in Section III suggests that speech scale functions have this property of utterance-independence; i.e., they are stable with respect to speech content.

This is not surprising for the following reason. We know that speech is composed of a small number of units (e.g., phonemes) that occur repeatedly with certain frequencies. Therefore, two sufficiently large samples of speech are likely to produce the same distribution of cepstral velocities in each cepstral neighborhood. Because the metric reflects the statistical distribution of those velocities (i.e., the velocity covariance matrix), two speech samples will lead to the same metric and the same scale function.

## III. EXPERIMENTAL RESULTS

We performed experiments on data from three speakers of American English, who were part of the Air Travel Information System (ATIS0) corpus of speaker-dependent training data[21]. As shown in Table I, these subjects were from a variety of accent regions and included males and females of various ages. The ATIS0 speech samples were recorded with a Sennheiser microphone at a 16 kHz sampling rate with 16 bits of depth. For each speaker, the clean speech sample was comprised of the unmodified data representing 11 or 12 sentences (approximately 80 s) of this corpus. Non-overlapping sets of sentences were used to define the clean speech samples of different speakers. The acoustic waveform of each sentence was Fourier transformed, after it had been Hamming-windowed in 24 ms time frames at 4 ms intervals. Each frame's power spectrum was used to compute 20 mel frequency cepstral coefficients[22] (MFCC). For each speaker, the set of sentences defined a time series of approximately $2 \times 10^4$ cepstra, which formed a trajectory in cepstral space. This trajectory densely traversed and retraversed a compact "speech domain", whose location, size, and shape depended on the speaker and channel characteristics (Fig. 2). The speech trajectory was dimensionally reduced by retaining its first two principal components, which contained approximately 95% of the data's variance. In Section IV, we propose to include more of the data's variance by retaining more principal components or by using more efficient non-linear methods of dimensional reduction.

Each trajectory was covered with a uniform 64 x 64 array of rectangular neighborhoods within which the clean speech metric was computed by the formula in Section II. If more data were available, it would be possible to increase the size of this array and thereby achieve higher "cepstral" resolution in our estimate of the metric. Then, parallel transport was defined in terms of an affine connection, which was given by the standard combination of the inverse metric and the metric's derivative in the Christoffel bracket[20]. For each speaker, we manually identified a tight cluster of trajectory segments that represented brief sounds in the clean speech sample (duration of each sound being 4 ms). These were used to determine a reference point and reference vectors ($x_0$ and $h_a$) that defined the origin and local orientation of the axes of the clean speech scale. Then, the complete scale (Fig. 2) was formed by parallel transporting these reference vectors away from the origin, as described in Section II. Scale values in regions immediately outside the traversed speech domain were estimated by extrapolating the scale values found by parallel transport within the speech region.

For each speaker, a corrupted speech sample was created from 11 or 12 *different* sentences by convolving each ATIS0 signal with a channel impulse response function and adding Gaussian white noise in the time domain. *Note that no sentence of* *the ATIS0 corpus was used twice for the same speaker or for different speakers*. Each speaker's speech was corrupted by one of two impulse responses (Fig. 3), which were synthesized by the "image source" method[23]. One of these functions described a relatively reverberant small room (reflectivity ~ 0.9), in which the speaker and microphone were 25 cm apart. The other impulse response corresponded to a "softer" version of the same room (reflectivity ~ 0.7), in which the speaker and microphone were 112 cm apart. Each impulse response included all reverberations with echo times less than 64 ms. After addition of noise, the SNR of the corrupted speech was 16-20 dB in each case. As above, the acoustic waveform of the corrupted speech was used to compute an MFCC time series, which formed a trajectory in cepstral space (Fig 4). This data was dimensionally reduced by retaining its first two principal components (containing approximately 89% of the data’s variance), and the metric and affine connection of corrupted speech were computed. Corrupted versions of the clean speech reference sounds were used to determine the corrupted reference information ($x'_0$ and $h'_a$), and the corrupted speech scale was then defined by parallel transporting these reference vectors away from the origin (Fig. 4). It is important to note that these brief reference sounds were the only information that was common to the derivations of the clean and corrupted speech scales, which were otherwise based on entirely different sets of sentences. In Section IV, we propose to use the methods of reference 12 to automatically derive reference vectors, thereby reducing the shared information to a single reference sound necessary to fix the location of each scale's origin. Notice that the scale function in Fig. 4 is ill-defined in the lower half of the speech domain. This is because of the relative paucity of data there, as well as the “edge effects” that we propose to remedy as described in Section IV.

Next, the scales of clean and corrupted speech were used to perform the channel conversion process described in Section II (Fig. 1). Specifically, the MFCCs of corrupted sentences were used to predict the MFCCs of clean versions of those sentences. First, the corrupted MFCCs were rescaled with the scale function of corrupted speech. The rescaled values were then mapped through the inverse scale function of clean speech to predict the MFCCs of the clean versions of the corrupted utterances. These were compared to the MFCCs of the actual clean versions of those utterances (i.e., the original ATIS0 versions before corruption by the channel's impulse response and noise). The upper panel of Fig. 5 shows an example of this type of comparison for the words "and make", spoken by speaker BF. Notice that the channel-converted MFCCs and the clean MFCCs were much closer to one another than were the corrupted and clean MFCCs after normalization by CMN. This result was produced by a procedure that does not involve the variation of any free parameters in order to best fit the data. The lower panel of Fig. 5 shows the distributions of Euclidean distances between the corrupted and clean MFCCs (after CMN) and between the channel-converted and clean MFCCs, at 1430 time points during all words in three typical sentences. These histograms (as well as the confidence intervals of their means in Table II) show that the channel conversion process did a much better job than CMN in moving the corrupted MFCCs close to the clean MFCCs at the great majority of time points. Furthermore, the new channel conversion procedure was comparable to the combination of CMN + SS in its ability to normalize speech from different channels.

Specifically, Fig. 5 and Table II show that the distribution of distances between the corrupted and clean MFCCs after channel conversion was comparable to the distribution of distances between corrupted and clean MFCCs, after "normalization" by CMN + SS. This is true despite the fact that the channel conversion procedure did not involve the measurement of noise levels required by spectral subtraction. Figures 6-7 show that similar results were obtained for the other speakers.

Two technical comments should be made at this point. First, recall that the scales of clean and corrupted speech were derived from dimensionally-reduced data. Therefore, the channel conversion process is only expected to predict the *dimensionally-reduced* MFCCs of clean versions of corrupted speech. It is NOT capable of predicting higher principal components of these MFCCs. Therefore, in Figs. 5-7, we compared how well the channel conversion process and conventional normalization methods (CMN alone or CMN + SS) could predict the dimensionally-reduced clean MFCCs from dimensionally-reduced corrupted MFCCs. However, similar results were obtained when we compared how well each method predicted the fully-dimensional clean MFCCs. For example, for speaker BF, the distance between fully-dimensional clean MFCCs and the corrupted MFCCs after channel conversion was equal to $29.5 \pm 0.9$, which is less than the distance between the fully-dimensional clean and corrupted MFCCs after CMN, namely $37.4 \pm 0.7$ (99% confidence intervals). Similar results were found for the other speakers.

Another technical issue concerns the ranges of the scale functions derived from the clean and corrupted speech samples. Each of these scale functions sweeps out a range of rescaled cepstra (*s* values) over the domain of the unrescaled cepstra (*x* values) of the corresponding speech sample. In principle, these two ranges should be the same, but they differed somewhat in actual practice. Because of this, some cepstra near the edges of the corrupted speech domain (constituting approximately 20% of the total) were rescaled to values outside the range of the clean speech scale function. Therefore, they fell outside the domain of the inverse of the clean speech scale function, and they could not be mapped through that inverse in order to compute their channel-converted values (Fig. 1). In Section IV, we propose to solve this problem by improved sampling of the speech data near the edges of the clean and corrupted speech domains.

## IV. CONCLUSIONS

Previous publications[12-16] described a new method of representing signal time series that essentially “filters out” the effects of unknown distortions. In this paper, the method was used to blindly create relatively channel-independent representations of speech cepstra. The experimental results suggest that the new technique is more successful than CMN and comparable to CMN + SS in its ability to decrease the signal’s channel dependence. Even better results can be expected if more of the data’s variance is retained in the dimensional reduction step and if longer speech samples are used to compute the metric and scale. Notice that the new method has the following advantages compared to conventional approaches to channel normalization: 1) it does not require prospective measurements of the channel's impulse response and noise; 2) it is a pure front end technology and avoids the computational demands of modifying or retraining the system's recognizer. In principle, an ASR system with the new front end can be trained in one environment and then used in another without additional measurements or retraining (D. N. Levin, patents pending). Of course, this hypothesis must be tested by comparing the word error rates of ASR systems with and without the new front end.

The implementation of these ideas can be improved in several ways:

*a) More accurate dimensional reduction.* Others have demonstrated that speech data have 3-5 underlying degrees of freedom, which presumably correspond to independent ways of moving the tongue, lips, soft palate, and other vocal structures[18-19]. However, for computational simplicity, the experiments in Section III were performed on speech signals that were approximated by their first two principal components, which contained 89-95% of the data's variance. This approximation obviously limited the accuracy of the attempted channel conversion by making it impossible to predict the higher principal components of the clean speech corresponding to observed corrupted speech. In addition, it probably violated the requirement that the rescaling method be applied to clean and corrupted speech data that are invertibly related. This is because the dimensionally-reduced cepstra need not have been related by an invertible transformation even though it is likely that the exact cepstra were so related (Section II). These observations suggest that better channel normalization can be expected if more of the data’s variance is included in the analysis. This can be done most simply by retaining 3-5 principal components of the data. Alternatively, one can use known methods of non-linear dimensional reduction[24-25], which find curved subspaces that optimally contain the data. Because of their greater generality, these techniques can be expected to find subspaces that contain more of the data's variance than linear subspaces of the same dimensionality.

*b) More accurate sampling of the cepstral trajectory.* As shown in Fig. 2, speech cepstra densely traverse and retraverse a compact domain that has a speaker-dependent and channel-dependent configuration. At the edges of this domain, the velocity of the speech trajectory tends to change its direction and magnitude as the trajectory curves in order to stay within the domain. However, the metric computation in Section III assumed constant velocity of the speech trajectory between the cepstral points corresponding to individual frames in the time domain. Therefore, the metric computation was likely to be less reliable at the edges of the traversed domain. This may explain why some rescaled values of cepstra near the edges of the corrupted speech domain fell outside the range of rescaled values of the clean speech sample, making it impossible to estimate the corresponding clean speech cepstra at those time points. This problem can be ameliorated in two ways. First, when a trajectory is near the edge of the cepstral domain of speech, it can be sampled at higher temporal resolution by computing cepstra corresponding to more closely spaced frames. Alternatively, the trajectory's velocity in these regions can be monitored, and those time intervals with rapidly changing velocity can be excluded from the metric computation. These steps are expected to increase the accuracy of the metric computation near the edges of the speech domain and thereby make it possible to channel-convert the signal in these regions.

*c) Automatic computation of most reference information.* In Section III, channel normalization was achieved by analyzing different utterances that propagated from a single speaker though clean and corrupted channels. The resulting waveforms were unrelated except for brief signals that were known to represent the same speech sounds from the two channels. One pair of these "reference" signals was used to define the locations of the origins of the clean and corrupted speech scales, and the other reference signals were used to derive vectors defining unit increments along each scale's axes at its origin. However, this procedure can be simplified in the following manner. As before, a single pair of reference signals can be used to establish the origins of the clean and corrupted speech scales. Then, the method demonstrated in reference 12 can be used to *automatically* derive vectors from the local directionality of the cepstral velocity distribution at each origin. These vectors can be used to define unit increments along each scale's axes at its origin. Once this procedure is in place, channel normalization will only depend on the identification of a single brief reference sound in each channel, analogous to the reference tone that a choir leader uses to coordinate the musical scales of individual singers prior to a concert.

*d) Adaptive channel normalization.* The system can adapt to changing channel conditions by using the most recent sample of corrupted speech in order to periodically update the metric and scale function. The updated scale of corrupted speech, together with the static scale of the clean speech, can be used to estimate the clean speech cepstra corresponding to observed corrupted speech. This adaptive rescaling technique was demonstrated successfully on one-dimensional signals in reference 13.

Of course, the reference signal of corrupted speech must also be periodically recomputed in order to update the origin of the corrupted speech scale (see item *c* above). This can be done automatically by having the system continuously track and identify the reference sound as it recurs in the corrupted speech. For example, suppose that the reference point is chosen to be the average cepstrum of a frequently-heard vowel-like sound with nearly stationary spectra. If the channel conditions change smoothly, it should be possible to track and identify this reference sound (possibly with the help of the recognizer) in order to update the origin of the corrupted speech scale.

It should be pointed out that the ideas in this paper can be applied in more general circumstances. In Section II, the embedding theorem was used to argue that the power spectrum of an acoustic channel's input and the MFCCs of its output are related by an invertible transformation, which characterizes the channel. By similar reasoning, the input power spectrum is invertibly related to the values of *any* set of spectral parameters that are used to characterize the output power, as long as those parameters are sufficiently numerous (more than 6-10) and as long as they average the power over many frequencies. Therefore, if a channel's output is detected with two different spectral parameter measurements (e.g., MFCCs vs. linear frequency cepstral coefficients), the pair of output time series will be invertibly related to one another, because each of them is invertibly related to the same time series of input power spectra. It follows that these output time series will rescale to the same form. This means that two ASR systems will derive the same rescaled representation of a signal, even though they used different spectral parameters to "sense" it and even though they received it through two different channels. In reference 13, this insensitivity of the rescaled signal to the choice of spectral measurements was demonstrated in experiments on bird songs having one underlying degree of freedom. The embedding theorem further guarantees that the power spectrum of the channel's input is a one-to-one function of the 3-5 parameters that characterize the instantaneous configuration of the speaker's vocal tract. It follows that the measured spectral parameters of the channel's output are also invertibly related to the speaker's vocal tract parameters. Because an invertible mapping is mathematically equivalent to a change of coordinate systems, the measured spectral parameters can be considered to describe the speaker's vocal tract configuration in a particular coordinate system. Therefore, if two ASR systems are "listening" to a given speaker through different channels and/or are measuring different spectral parameters, they are both recording the trajectory of the speaker's vocal tract, although they are describing it in different coordinate systems. Mathematically speaking, the "inner" properties of a geometrical figure are those that are independent of the coordinate system used to numerically describe it (or, equivalently, independent of transformations of the figure in a fixed coordinate system). Geometry seeks to find these "inner" properties and is less concerned with the figure's "outer" properties: i.e., aspects of its description that depend on the choice of coordinate system. For example, Euclidean geometry studies the properties of a figure that don't depend on how the coordinate system has been rotated or translated, and differential geometry focuses on properties that are invariant under general, non-linear coordinate transformations. From a geometrical perspective, the rescaled form of a speech signal is an "inner" property of the vocal tract motion because it gives a coordinate-independent description of that motion. In a sense, the details of the detection process (e.g., the nature of the channel and measured spectral parameters) only influence the appearance (i.e., the "outer" aspect) of the articulatory gesture, which does not affect the rescaled signal.

It is interesting to consider the possibility that rescaling can be used to create speaker-independent representations of speech signals. This idea is outlined here, but in other reports this process was experimentally demonstrated with synthetic speech-like sounds having a single degree of freedom[13, 16]. Suppose there is an invertible transformation that consistently maps the instantaneous configuration of one speaker's vocal tract onto the configuration of the other speaker's vocal tract when they utter the same words. This is equivalent to the assumption that the two speakers' articulatory gestures consistently mimic each other when the same words are uttered. In this case, the MFCC trajectories of the two speakers' signals must be invertibly related to one another, because each is invertibly related to the trajectory of the originating vocal tract and the configurations of the two vocal tracts are invertibly related to each other. It follows that the MFCC trajectories of the two signals have the same rescaled form; i.e., a speaker-independent form. This will be true even if their speaking rates differ by a multiplicative constant because the parallel transport process is independent of temporal scaling, as noted in Section II. Notice that the spectral parameter trajectories of both speakers' signals can be considered to describe the vocal tract trajectory of one of the speakers in two different coordinate systems. More generally, *all* of the different spectral parameter trajectories of a given utterance (corresponding to different combinations of speakers, channels, and spectral measurements) can be generated by describing the vocal tract trajectory of a single speaker in different coordinate systems. From this point of view, the rescaled form of an utterance's trajectory is expected to be independent of the speaker, channel, and detection process because the rescaled signal is sensitive to the "inner" aspects of that vocal trajectory, not to the choice of coordinate system used to describe it.

The rescaling procedure may be useful for addressing the problem of speaker identification, because it cleanly separates speech content from the characteristics of speaker and channel. Specifically, each speaker/channel combination is associated with a non-linear scale function or coordinate system that covers the patch of spectral parameter space traversed by the signals from that source (e.g., the warped grid of *s* isoclines in Figs. 2 and 4). The speaker/channel is characterized by the location and configuration of this scale. The content of an utterance is given by describing its spectral parameter trajectory in this coordinate system, which can be considered to define a "medium" on which the message is "written". Although the trajectory of a given utterance in the MFCC coordinate system is translated and warped in a channel-dependent and speaker-dependent manner, in the special coordinate system associated with its source, it is rescaled to a speaker-independent and channel-independent form because that coordinate system is translated and warped in the same way.

| SPEAKER | GENDER | AGE | ACCENT |
|---|---|---|---|
| BF | male | 20 | western |
| B0 | female | 40 | north midland |
| B5 | female | 30 | south midland |

**Table I.** Characteristics of the subjects chosen from the ATIS0 database (21).

| SPEAKER | REVERBERATIONS | SNR | CMN | CMN + SS | CHANNEL CONVERSION |
|---|---|---|---|---|---|
| BF | Fig. 3a | 16 dB | 35.4±0.9 | 22.9±0.6 | 23.4±1.0 |
| B0 | Fig. 3b | 16 dB | 40.1±0.8 | 28.4±0.7 | 27.8±1.3 |
| B5 | Fig. 3a | 20 dB | 53.8±0.6 | 36.7±0.4 | 35.5±1.1 |

**Table II.** The mean Euclidean distance between clean cepstra and those corrupted by reverberations and noise, after "normalization" by CMN, CMN + SS, and the new channel conversion procedure. In each case, the cepstra describe 1430-2860 time points in all words in three typical sentences. The 99% confidence interval of each mean distance is listed.

## Figure Captions

**Figure 1.** Schematic outline of the new method. a) The cepstral trajectory of an utterance from channel #1. b) The scale function derived from a speech sample from channel #1. a') The cepstral trajectory of the channel #2 version of the utterance in *a*. b') The scale function derived from a channel #2 speech sample. c) The trajectory found by using *b* to rescale *a*, which is also equal to the trajectory found by using *b'* to rescale *a'*. The dotted arrow shows how the channel #1 cepstra (*a*) can be converted into the channel #2 cepstra (*a'*) by mapping the rescaled values of *a* through the inverse of the channel #2 scale function (*b'*).

**Figure 2.** Left: The trajectory of the first two principal components of the cepstra of 12 clean sentences from speaker BF. This figure has been rotated and rescaled along each axis to show detail. Right: The scale function derived from the left panel. The thin black (thick gray) lines are $s_2$ ($s_1$) isoclines.

**Figure 3.** Impulse response functions of a "hard" room with a close (25 cm) microphone (*a*) and a "soft" room with a distant (112 cm) microphone (*b*). The time axis is labeled by the number of 16 kHz samples. The impulse response at $t=0$ is unity.

**Figure 4.** Left: The cepstral trajectory of 12 sentences from speaker BF, after corruption with reverberations (Fig. 3a) and noise. This figure has been rotated and rescaled along each axis to show detail. Right: The scale function derived from the left panel. The thin black (thick gray) lines are $s_2$ ($s_1$) isoclines.

**Figure 5**. Speaker BF. Upper: The dark solid and dashed lines show the MFCCs of the clean and corrupted versions of the words "and make", respectively, after "normalization" by CMN. The solid gray line shows the corrupted MFCCs after the new

channel conversion procedure. Lower: The distribution of Euclidean distances between the corrupted and clean cepstra after CMN (left), after CMN + SS (center), and after the new channel conversion process (right). These distributions describe cepstra at 1430 time points during all words in three typical sentences.

**Figure 6**. Speaker B0. Upper: The dark solid and dashed lines show the MFCCs of the clean and corrupted versions of the word “flight”, respectively, after “normalization” by CMN. The solid gray line shows the corrupted MFCCs after the new channel conversion procedure. Lower: The distribution of Euclidean distances between the corrupted and clean cepstra after CMN (left), after CMN + SS (center), and after the new channel conversion process (right). These distributions describe cepstra at 1863 time points during all words in three typical sentences.

**Figure 7**. Speaker B5. Upper: The dark solid and dashed lines show the MFCCs of the clean and corrupted versions of the word “airlines”, respectively, after “normalization” by CMN. The solid gray line shows the corrupted MFCCs after the new channel conversion procedure. Lower: The distribution of Euclidean distances between the corrupted and clean cepstra after CMN (left), after CMN + SS (center), and after the new channel conversion process (right). These distributions describe cepstra at 2860 time points during all words in three typical sentences.

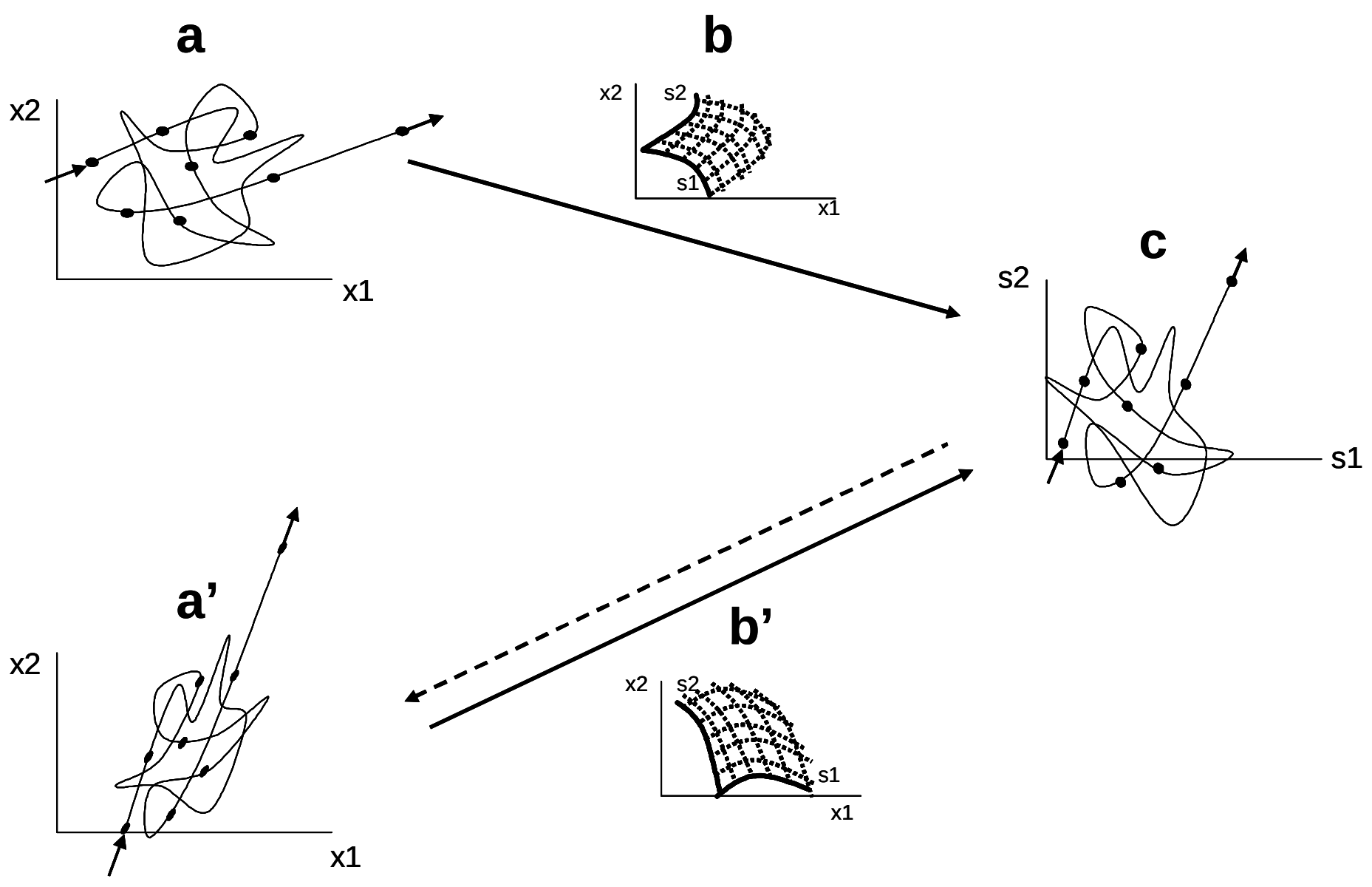

**Figure 1.** Schematic outline of the new method. a) The cepstral trajectory of an utterance from channel #1. b) The scale function derived from a speech sample from channel #1. a’) The cepstral trajectory of the channel #2 version of the utterance in *a*. b’) The scale function derived from a channel #2 speech sample. c) The trajectory found by using *b* to rescale *a*, which is also equal to the trajectory found by using *b’* to rescale *a’*. The dotted arrow shows how the channel #1 cepstra (*a*) can be converted into the channel #2 cepstra (*a’*) by mapping the rescaled values of *a* through the inverse of the channel #2 scale function (*b’*).

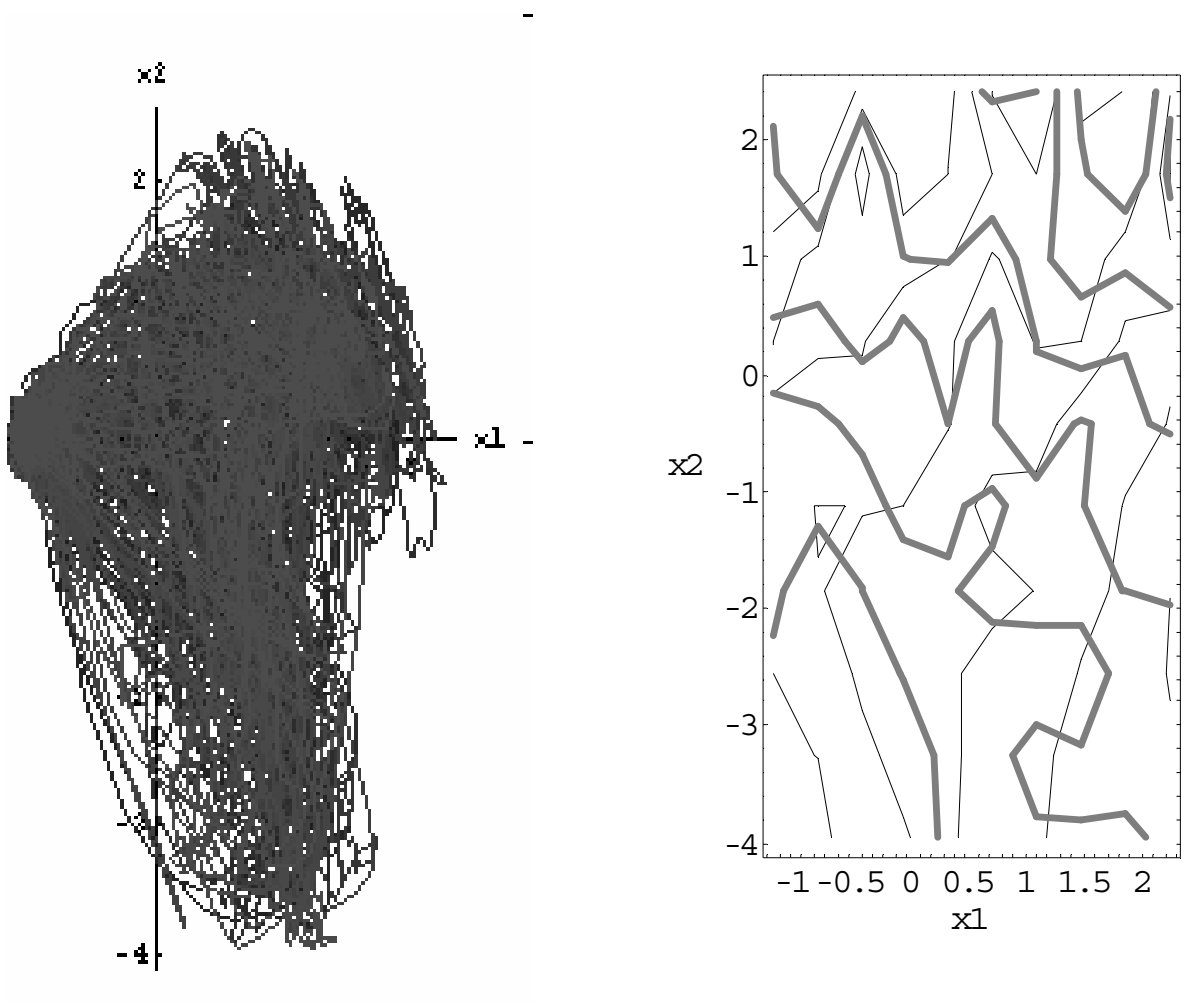

**Figure 2.** Left: The trajectory of the first two principal components of the cepstra of 12 clean sentences from speaker BF. This figure has been rotated and rescaled along each axis to show detail. Right: The scale function derived from the left panel. The thin black (thick gray) lines are $s_2$ ($s_1$) isoclines.

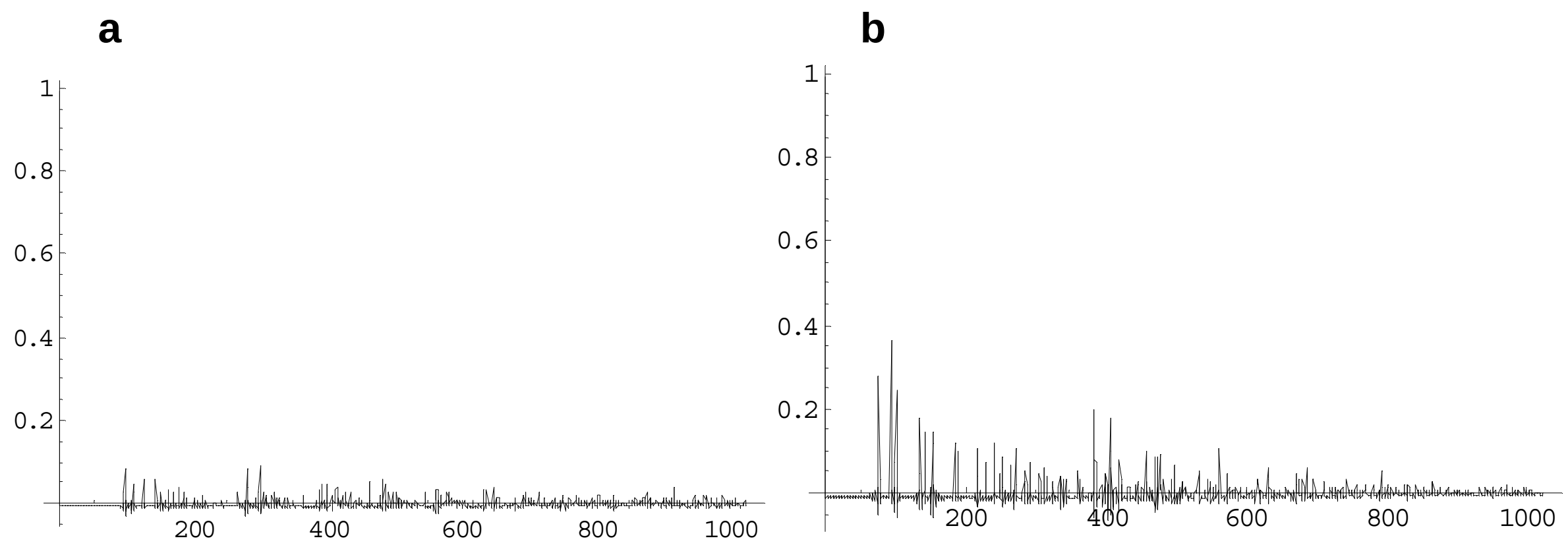

**Figure 3.** Impulse response functions of a “hard” room with a close (25 cm) microphone (*a*) and a “soft” room with a distant (112 cm) microphone (*b*). The time axis is labeled by the number of 16 kHz samples. The impulse response at *t=0* is unity.

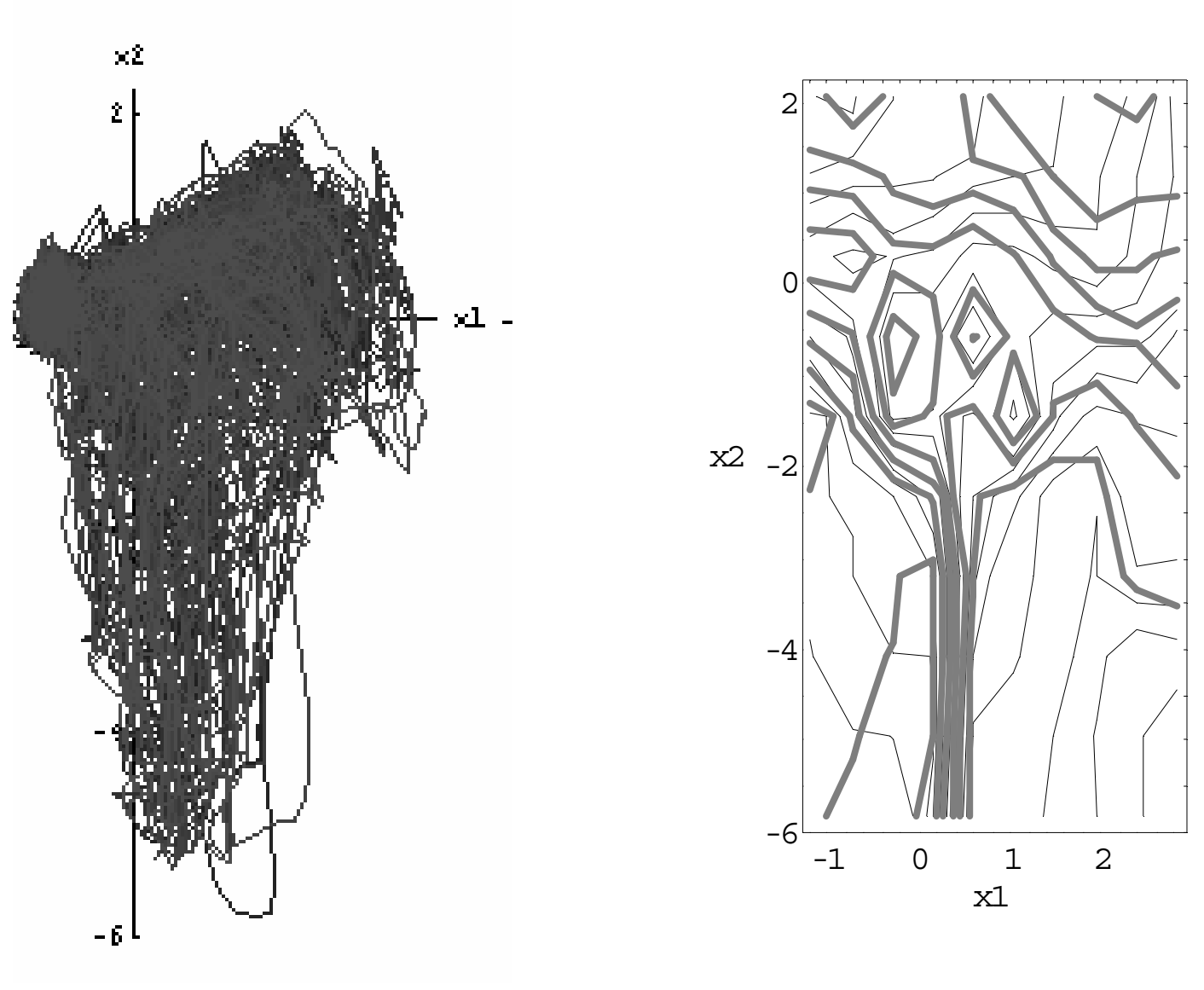

**Figure 4.** Left: The cepstral trajectory of 12 sentences from speaker BF, after corruption with reverberations (Fig. 3a) and noise. This figure has been rotated and rescaled along each axis to show detail. Right: The scale function derived from the left panel. The thin black (thick gray) lines are $s_2$ ($s_1$) isoclines.

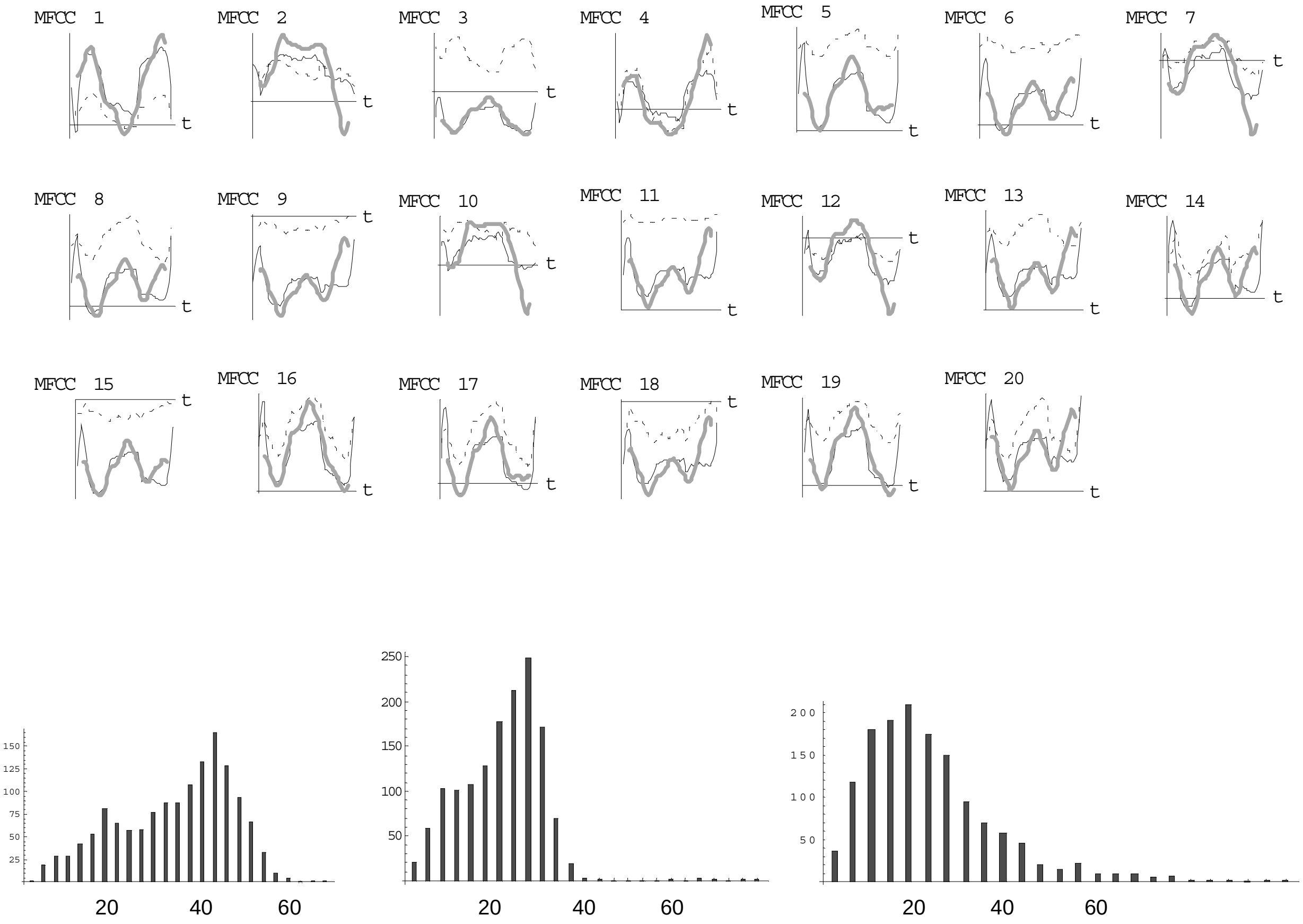

**Figure 5**. Speaker BF. Upper: The dark solid and dashed lines show the MFCCs of the clean and corrupted versions of the words “and make”, respectively, after “normalization” by CMN. The solid gray line shows the corrupted MFCCs after the new channel conversion procedure. Lower: The distribution of Euclidean distances between the corrupted and clean cepstra after CMN (left), after CMN + SS (center), and after the new channel conversion process (right). These distributions describe cepstra at 1430 time points during all words in three typical sentences.

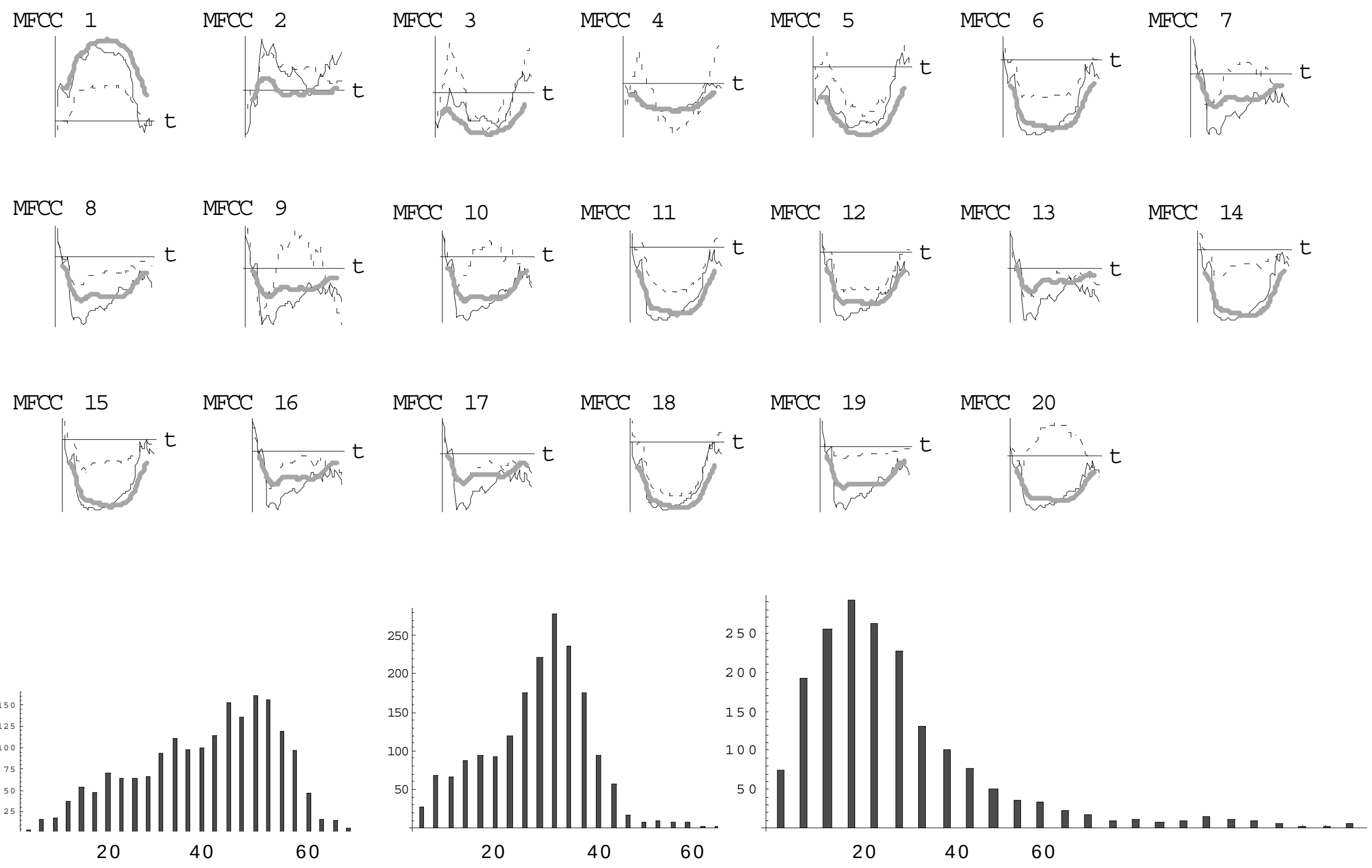

**Figure 6**. Speaker B0. Upper: The dark solid and dashed lines show the MFCCs of the clean and corrupted versions of the word “flight”, respectively, after “normalization” by CMN. The solid gray line shows the corrupted MFCCs after the new channel conversion procedure. Lower: The distribution of Euclidean distances between the corrupted and clean cepstra after CMN (left), after CMN + SS (center), and after the new channel conversion process (right). These distributions describe cepstra at 1863 time points during all words in three typical sentences.

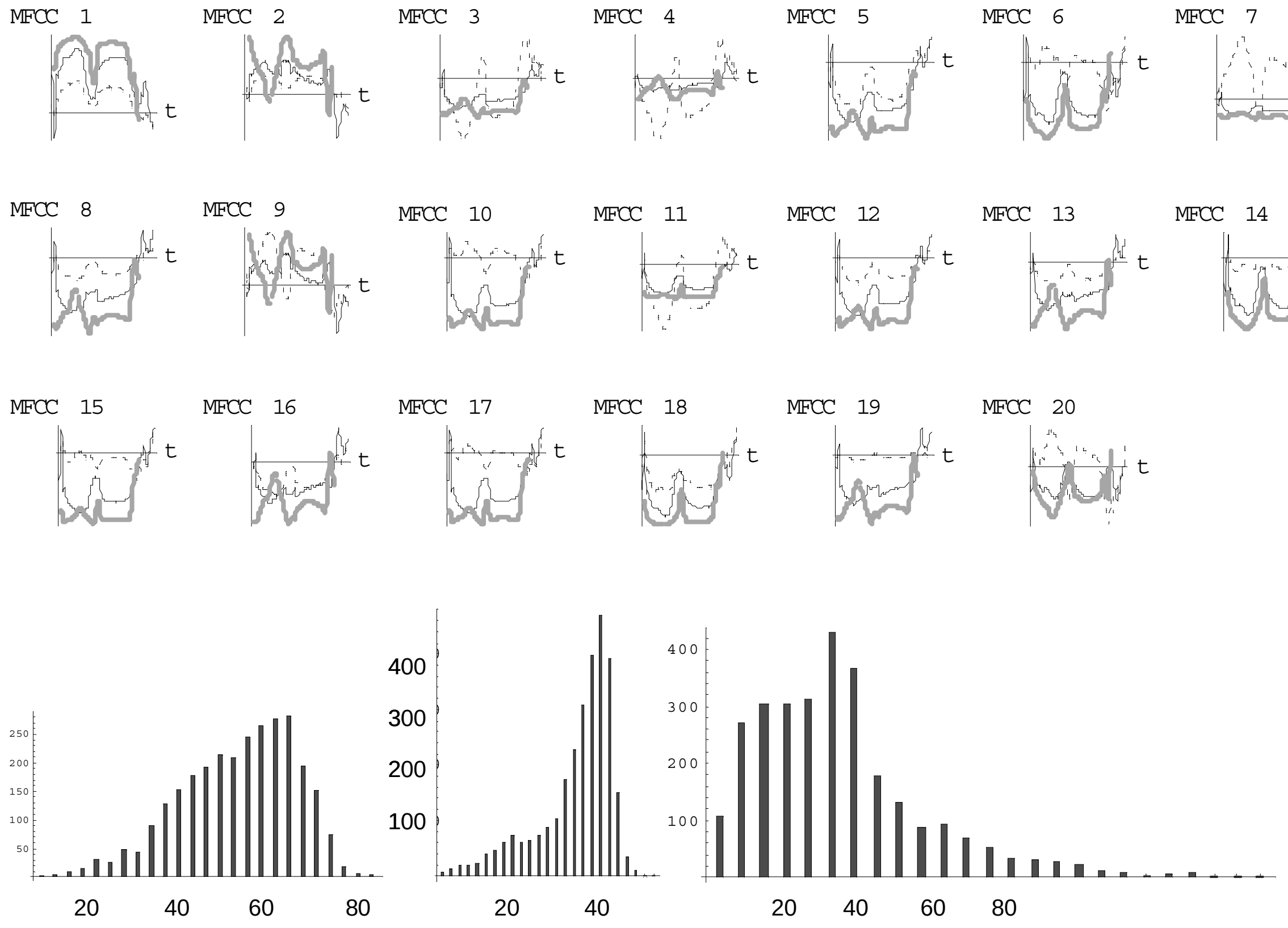

**Figure 7**. Speaker B5. Upper: The dark solid and dashed lines show the MFCCs of the clean and corrupted versions of the word "airlines", respectively, after "normalization" by CMN. The solid gray line shows the corrupted MFCCs after the new channel conversion procedure. Lower: The distribution of Euclidean distances between the corrupted and clean cepstra after CMN (left), after CMN + SS (center), and after the new channel conversion process (right). These distributions describe cepstra at 2860 time points during all words in three typical sentences.